# INTEGRATING EXTRACTED INFORMATION FROM BERT AND MULTIPLE EMBEDDING METHODS WITH THE DEEP NEURAL NETWORK FOR HUMOUR DETECTION

Rida Miraj and Masaki Aono

Department of Computer Science and Engineering, Toyohashi University of Technology, Aichi, Japan

*ABSTRACT*

*Humour detection from sentences has been an interesting and challenging task in the last few years. In attempts to highlight humour detection, most research was conducted using traditional approaches of embedding, e.g., Word2Vec or Glove. Recently BERT sentence embedding has also been used for this task. In this paper, we propose a framework for humour detection in short texts taken from news headlines. Our proposed framework (IBEN) attempts to extract information from written text via the use of different layers of BERT. After several trials, weights were assigned to different layers of the BERT model. The extracted information was then sent to a Bi-GRU neural network as an embedding matrix. We utilized the properties of some external embedding models. A multi-kernel convolution in our neural network was also employed to extract higher-level sentence representations. This framework performed very well on the task of humour detection.*

## 1. INTRODUCTION

Humour is a way of giving enjoyment and provoking laughter. The creation and perception of humour has received a lot of attention in recent years. According to some scholars, humour leads towards the enhancement of the human's health and mood [1]–[4]. With the passage of time, a lot of work has been done in the field of Artificial Intelligent (AI) field toward simulating human intelligence in machines or computer systems [5]. Natural Language Processing (NLP), a branch of AI, is used to understand human languages despite their diversity and complexity, which are open challenges for NLP systems and communities [6], [7].

Table 1.  Some headlines from dataset and their edits, mean funniness grades.

| ID | Original Headlines | Substitute | Grade |
|---|---|---|---|
| 1 | Trump wants you to take his tweets seriously. His aides don't | hair | 2.8 |
| 2 | 4 arrested in Sydney raids to stop terrorist attack | kangaroo | 2.4 |
| 3 | 4 soldiers killed in Nagorno-Karabakh fighting: Officials | rabbits | 0.0 |

What makes us laugh when reading a funny sentence? Mostly it depends on the cultural area and background of a person. Humour cannot be defined in standards as it varies with nation to nation, area to area, people to people. In other words, humour is universal, the different cultures see the humour in various ways [8]. Many types of humour require substantial external knowledge such as irony, wordplay, metaphor, and sarcasm. These factors make the task of humour recognition difficult. Recently, with the advance of deep learning that allows end-to-end training with big data without human intervention of feature selection, humour recognition becomes promising. The recognition of humour in text has been receiving much attention [9], [10].

The detection of humour from a small and formal sentence is a unique challenge for the research community. To address the challenge of humour detection, Hossain et al. [10] presented a task that focuses on detecting humour in English news headlines with micro-edits. Micro change was defined as one of the following alternatives: entity→noun, noun→noun and verb→verb. Each edited headline was scored by five judges according to [10], each of whom assigned a grade from 0 (not funny) to 3 (funny). The funniness level of each headline is the mean of its five funniness grades. Example of sample from the data is presented in Table 1. The goal is to determine how machines can understand humour generated by short edits. The edited headlines are named as humicroedit. Accurate scoring of the funniness from micro-edits can serve as a footstone of humorous text generation [10]. Figure 1 depicts how a single word is replaced with another word to make the news headline of reddit website funny.

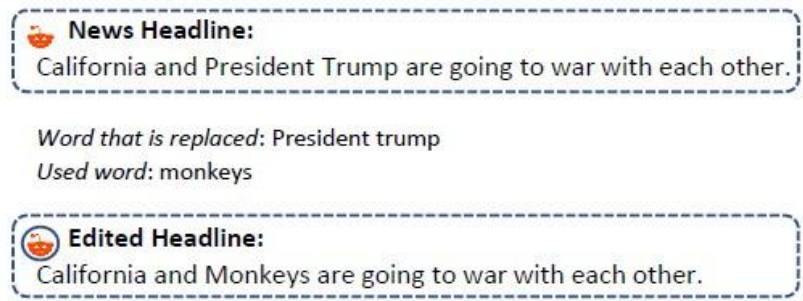

Figure 1. Example of humicroedit

The rest of the paper is structured as follows: Section 2 provides a brief rundown of prior research. In Section 3, we introduce our proposed humour detection framework. Section 3 includes experiments and evaluations. Section 4 contains some concluding remarks as well as prospective directions for our research.

## 2. RELATED WORK

Humour is a ubiquitous, elusive event that exists all around the world. In previous research and studies, mostly the problems related to humour were based on binary classification or based on the selection of linguistic features. Purandare and Litman analysed humorous spoken conversations as data from a classic comedy television and used standard supervised learning classifiers to identify humorous speech in the conversation [11]. Taylor and Mazlack used a methodology that was based on the extraction of structural patterns and the peculiar structure of jokes on newcite [12]. Luke de Oliveira and Alfredo applied recurrent neural network (RNN) and convolutional neural networks (CNNs) to humour detection from reviews in a Yelp dataset [13]. Some researchers in [2], [14]–[16] studied humour detection in several languages to classify each tweet into a joke or not. Jihang Mao and Wanli Liu classified the dataset into a binary classification that was obtained from Twitter in the Spanish language [2]. In [16], author predicted the level of funniness in tweets that was actually score value prediction based on the average value of 5-stars. To predict the funniness value, several machine and deep learning techniques have been used. In [17][17], a model that was consist of Bidirectional-LSTM (BiLSTM) and LSTM was used in the HAHA-2019 task. For humour recognition from texts, Vladislav Blinov used many classification algorithms, including the linear Support Vector Machine (SVM) [18]. Barbieri and Saggion [9] widely investigated features for automatically detecting irony and humour. [19] presented a large-scale annotated corpus of sarcasm and provided baseline systems for sarcasm detection.

However, most of the existing related approaches attempts to utilize the traditional neural network models to detect the humour from tweets. However, Hossain et al. [20] presented a slightly different type of challenge, namely, an attempt to investigate how small edits can turn a text from non-funny to funny. For this purpose, we have built a robust model, named IBEN (Integrating BERT and other Embeddings with Neural Network). We propose a framework that combines the inner layers information of BERT with Bi-GRU and uses multiple word embeddings with multikernel convolution, and Bi-GRU in unified architecture. Experimental results on edited news headlines demonstrate the efficacy of our framework.

## 3. FRAMEWORK

In this section, we describe the details of our proposed framework for humour detection. Figure 2 depicts an overview of our proposed framework. On one side, we utilize the BERT layers for word embedding purposes. The embedding matrix is then fed into the embedding layer of our neural network.

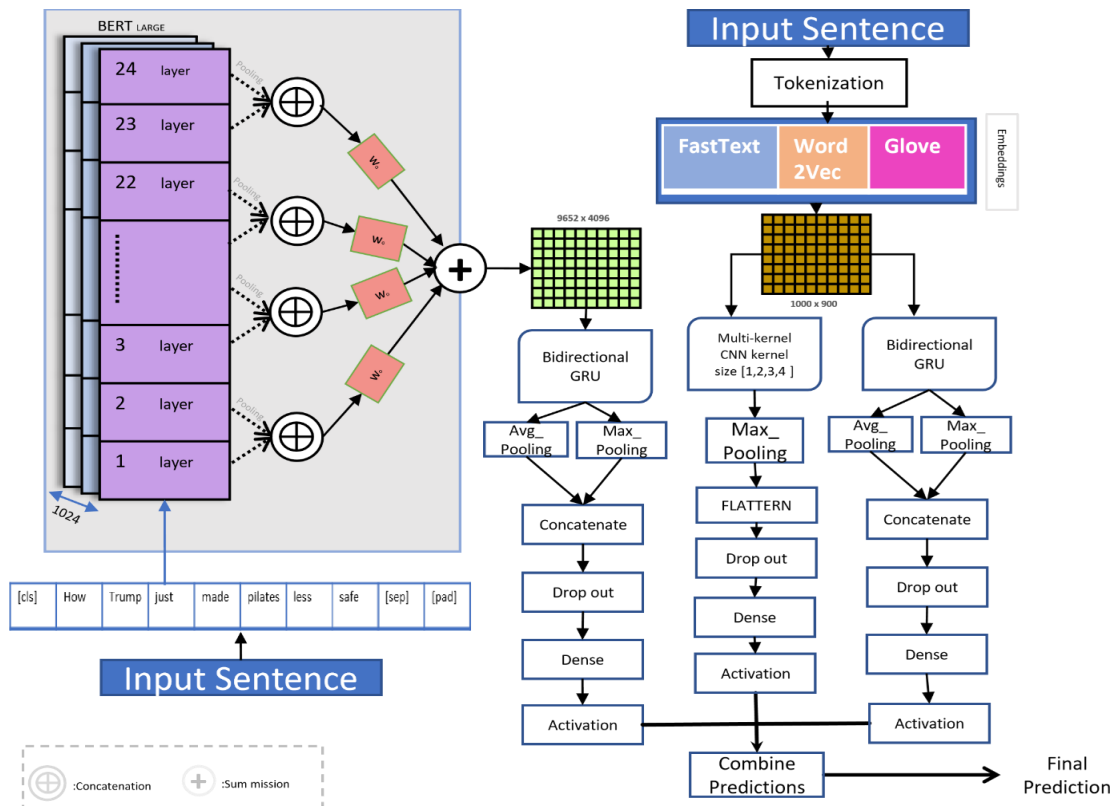

Figure 2. Proposed framework (IBEN)

The multi-kernel convolution filters, on the other hand, are used to remove higher-level feature sequences from the appended embeddings. After receiving the predictions from these modules, the results are blended and used to determine the degree of funniness. Next, we describe each component elaborately.

## 3.1. BERT Feature Extraction

BERT [21] is a recent language representation model that has accomplished reaching diverse language understanding benchmarks remarkably well, which indicates the possibility that BERT networks capture essential structural information about languages. BERT builds on Transformer networks [22] to pre-train bidirectional representations by conditioning on both left and right contexts jointly in all layers. The transformer network inside BERT uses the encoder which has a self-attention layer. Self-attention allows the current node to rely not only on the current term but also on the semantics of the context. Different BERT layers capture different information. Our target is to extract the hidden information denoted as {$h_L = h_1, h_2, h_3,…, h_{24}$} where L is the number of layers in the BERT model. For extraction, we use two pooling strategies. One is taking the average of the hidden state of the encoding layer. The second pooling technique is taking the maximum of the hidden state of the encoding layer. The reason to use these strategies is: In average pooling, the meaning of a sentence is represented by all words contained, and in max pooling, the meaning is represented by a small number of keywords and their salient features only. Two special tokens [CLS] and [SEP] are padded onto the beginning and the end of an input sequence, respectively. Our BERT model is only pre-trained and not fine-tuned for our task, embeddings from those two symbols are meaningless here. After the pooling, the extracted information is concatenated. This extraction method is applied on several layers of the BERT model because every layer has something informative inside of it. i.e., the last layer is closed to the training output, so it may give a biased representation towards the training targets. The first layer is closed to the word embedding, this may preserve the original word information (without self-attention) [23]. This trade-off between different layers can be beneficial in feature extraction for our task. The rest of the layers are processed accordingly. We can define the above process as follows:

$$h_L^o = AVG(h_L) \odot MAX(h_L),$$
$$h_{out}^L = (h_L^o) \odot (h_{L-1}^o),$$
$$E_L^o = \sum_{l=1}^{l} \alpha[h_{out}^L].$$

In the above equations, the sign $\odot$ is concatenation operation. In the next step, the summation $E^o_h$ of the concatenated layers are completed after adding weights to them as shown in figure 2.

## 3.2. Embeddings

Word embedding, a distributed representation of terms, is considered one of the most common representations of document vocabulary due to its capacity to capture a context of a word within a document as well as estimate semantic similarity and association with other words. Such a vector representation of documents can help learning algorithms to achieve better performances in various natural language processing (NLP) applications [22][24][25][26] [26] [24]–[26].

In prior work, target information for humour detection was gained from traditional methods of embedding. As shown in figure 2, we also use these embedding techniques in our proposed framework. We create a unified word vector matrix by concatenating the vector representations of the news to incorporate the target information. The dimensionality of the matrix $E_{glove,fasttext,word2vec}$ will be $L \times D$, where length L is the length, and D denotes the word-vector dimension. For obtaining the vector representation of words, we use a pre-trained word embedding model.

## 3.3. Bi-GRU

Recurrent Neural Network is widely used in the NLP field, which can learn context information of one word. Long Short-Term Memory is designed to solve the RNN gradient vanishing problem, especially learning long sentences [27]. Grate Recurrent Unit is a simplified LSTM cell structure [28]. Taking advantage of its simple cell, GRU can get a close performance to LSTM with less time. With the emerging trend of deep learning, recurrent networks are the most popular for sequential tasks. A Bidirectional GRU, or BiGRU, is a sequence processing model that consists of two GRUs; one taking the input in a forward direction, and the other in a backward direction as shown in figure 3. It is a bidirectional recurrent neural network with only the input and forget gates. Figure 4 shows the typical structure of gated recurrent unit where Z is the update gate and r is the reset gate [29].

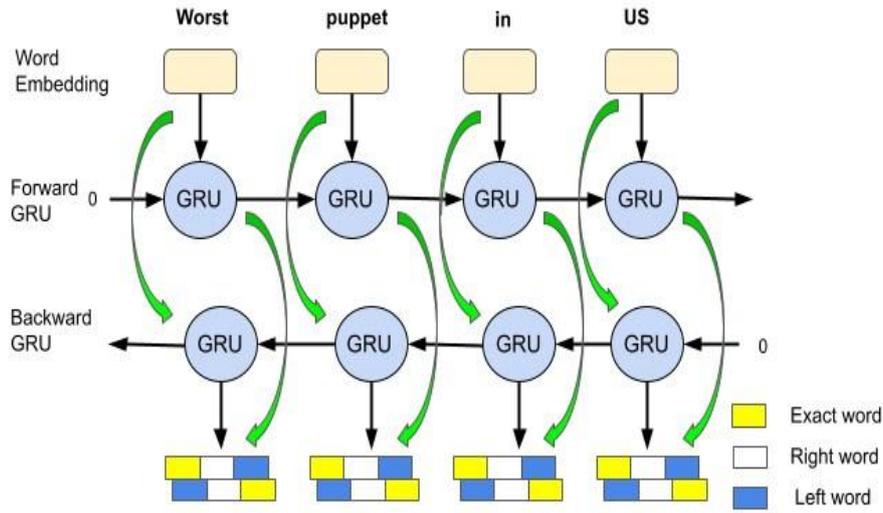

Figure 3. Bidirectional-GRU

The overall network (Right direction and Left direction) and its equations for GRU network is as follows.

Right direction: $\overrightarrow{h_t}^{(i)}$

$$\overrightarrow{z_t}^{(i)} = \partial(\overrightarrow{W}_{(i)}^{(z)} x_t^i + \overrightarrow{U}_{(i)}^{(r)} h_{t-1}^{(i)})$$
$$\overrightarrow{r_t}^{(i)} = \partial\left(\overrightarrow{W}_{(i)}^{(r)} x_t^i + \overrightarrow{U}_{(i)}^{(r)} h_{t-1}^{(i)}\right)$$
$$\overrightarrow{\tilde{h}}^{(i)} = \tanh(\overrightarrow{W}_{(i)} x_t + r_t^\circ \overrightarrow{U}_{(i)} h_{t-1})$$
$$\overrightarrow{h_t}^{(i)} = z_t^{(i)} \circ h_{t-1}^{(i)} + (1 - z_t^{(i)}) \circ \tilde{h}_{(t)}^{(i)} h_{t-1}^{(i)}$$

Left direction: $\overleftarrow{h_t}^{(i)}$

$$\overleftarrow{z_t}^{(i)} = \partial(\overleftarrow{W}_{(i)}^{(z)} x_t^i + \overleftarrow{U}_{(i)}^{(r)} h_{t-1}^{(i)})$$
$$\overleftarrow{r_t}^{(i)} = \partial(\overleftarrow{W}_{(i)}^{(r)} x_t^i + \overleftarrow{U}_{(i)}^{(r)} h_{t-1}^{(i)})$$
$$\overleftarrow{\tilde{h}}^{(i)} = \tanh(\overleftarrow{W}_{(i)} x_t + r_t^\circ \overleftarrow{U}_{(i)} h_{t-1})$$
$$\overleftarrow{h_t}^{(i)} = z_t^{(i)} \circ h_{t-1}^{(i)} + (1 - z_t^{(i)}) \circ \tilde{h}_{(t)}^{(i)} h_{t-1}^{(i)}$$

Where $z_t$ represents update gate, $r_t$ is reset gate, $\tilde{h}_t$ is the reset memory, and $h_t$ is new memory. GRUs use less training parameters and therefore use less memory, execute faster and train faster than LSTM's. In our proposed framework, we utilize the Bi-GRU model. The embeddings $E^o{}_h$ and $E_{glove, fasttext, word2vec}$ passes through the Bi-GRU layer separately. Then max & avg pooling functions are applied which are concatenated to form a feature vector.

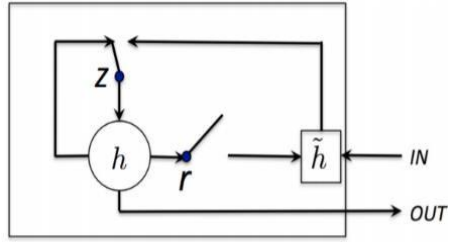

Figure 4. Typical structure of a GRU

## 3.4. Multi-Kernel Convolution

In our multi-kernel convolution, we adopt the idea proposed by [30] to extract higher-level features. The input of this module is the embedding matrix generated in the embedding layer. We then perform the convolution on it using a filter. We apply multiple convolutions based on four different kernel sizes, i.e., the size of the convolution filters: 1,2, 3, and 4. Each filter produces the corresponding feature maps after executing convolutions, and then a max-pooling function is used to generate a univariate feature vector. Finally, each kernel's feature vectors are concatenated to create a new high-level feature vector.

## 3.5. Humour Prediction and Model Training

We concatenate the results from the BERT based Bi-GRU model and Embedding based CNN & Bi-GRU after passing them through a fully connected linear layer for humour detection. We consider mean square error (mse) as the loss function and train the models by minimizing the error, which is defined as:

$$mse = \sum_{i=1}^{n} |y_i - \hat{y}_i|$$

where i is the sample index with its true label $y_i$. $\hat{y}_i$ is the estimated value. We use the stochastic gradient descent (SGD) to learn the model parameter and adopt the Adam optimizer [31].

## 4. EXPERIMENTS AND EVALUATIONS

### 4.1. Data Processing

To validate the effectiveness of our proposed framework for humour detection, we made use of a dataset used in the SemEval-2020 Task 7 [20]. In this section, we would like to point out some interesting facts about the used data. The training set consists of 9652 news headlines, the validation set, and the test set consist of 2419 and 3024 news headlines, respectively. The dataset was collected from the Reddit website for predicting the funniness score. The score ranges from 0 to 3, where 0 means not funny and 3 means funniest of all. For each example, the dataset offers annotation in the form of grades (0,1,2 and 3) of humour assessment in sorted descending order and the mean of these grades. In most cases, there are five grades per dataset sample (sometimes more). As can be seen from graphs in Figure 2, the dataset is imbalanced, and the high grades are rare. In the histogram, we can see the number of samples in the train set that have a mean grade in each bin. The bins are left-inclusive. Before training, the pre-processing and cleaning procedure has been done. In the first step, the original headlines are changed with the given edit words. Following that, many pre-processing packages were tried, however the model's best outcomes come from using the initial data with just a few amounts of pre-processing. For cleaning the data, existing pre-processing packages such as spaCy[32] and NLTK's [33] is used. Moreover, stopwords play a negative role because they do not carry any opinion-oriented information and may damage the performance of the classifiers. For stopword removal, we utilize the NLTK's [33] standard stoplist. Next, we need to pad the input. The reason behind padding the input is that text sentences have varying length, however models used in our framework expects input instances with the same length. Therefore, we need to convert our sentences into fixed-length vectors. For padding, the maximum sentence length that is set in our framework is 40. For the evaluation measure, root mean square error (rmse) is employed.

$$rmse = \sqrt{\frac{1}{n}\sum_{i=1}^{n}(y_i - \hat{y}_i)^2}$$

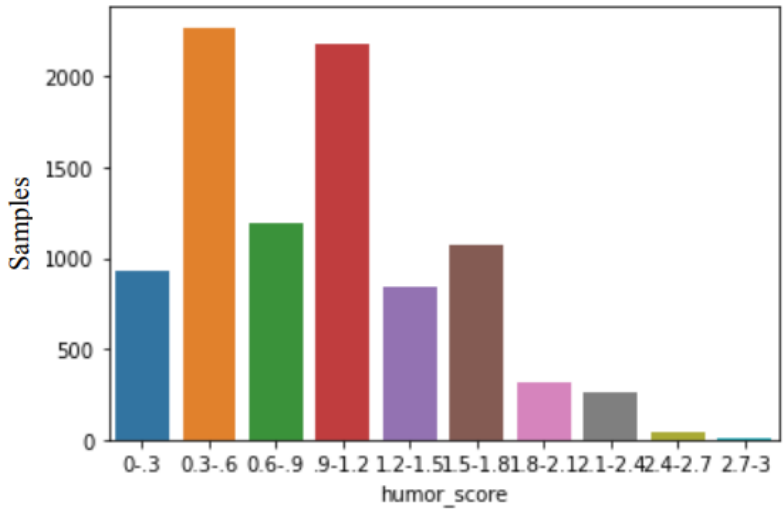

Figure 5. the number of samples with mean grade

## 4.2. Model Configuration

In the following, we describe the set of parameters that we used in our framework during experiments. We used three embedding models to initialize the word embeddings in the embedding layer. The embedding models are 300-dimensional FastText embedding model pretrained on Wikipedia with 16B tokens [34], 300-dimensional Glove embedding model with 2.2M vocab and 840B tokens [35] and 300-dimensional Word2Vec embedding model pre-trained on part of the Google News dataset [36].

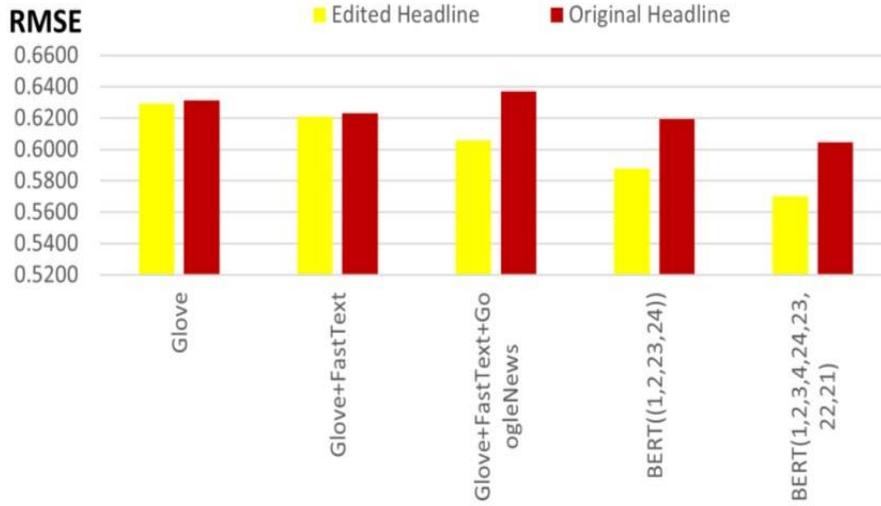

Figure 6. Performances of different models on Original and Edited headlines

We utilized four kernel sizes (1,2,3, and 4) for the multi-kernel convolution, with a total of 36 filters. We use Bert-as-Service [37] for extracting information from the BERT model. In our system, the layers of BERT-Large(uncased) are used in the following manner. $h_1^l \odot h_2^l, h_3^l \odot h_4^l, \ldots, h_{21}^l \odot h_{22}^l, h_{23}^l \odot h_{24}^l$. The framework which we used to design our model was based on TensorFlow [38] and training of our model is done on a GPU [39] to capture the benefit from the efficiency of parallel computation of tensors. We trained all models for a max of 25 epochs with a batch size of 16 and an initial learning rate of 0.001 via Adam optimizer. We recorded the findings based on these settings in this article. The other parameters were left at their default settings unless otherwise specified.

### 4.3. Results and Analysis

Our target is to detect the level of funniness from news headlines that are not supposed to be funny. Here, we used the dataset in two different manners to show the efficacy of our framework (IBEN). We showed some summarized experimental results of original and edited headlines in Table 2. At first, we reported the results based on a naive baseline system. Next, we reported the results of our proposed framework. In order to estimate the effect of each component of our framework, we showed the performances of each component individually. Figure 3 shows the comparison of each component on original and edited headlines. From the results, it can be observed that the simple embedding technique on original and edited headlines gave almost the same RMSE error, which shows it cannot significantly distinguish between being funny or not. For the validation dataset, these techniques produced biased results for most of the cases. Having in-depth knowledge of a sentence via BERT layers make it better regarding original and edited headlines. We did not perform multi-kernel convolution on BERT based Embedding due to large computational time.

Table 2. Comparative results with different experimental settings for original news headlines and for edited headlines. The best results are highlighted in boldface.

| EMBEDDING | Model | RMSE |
|---|---|---|
| **Baseline** | | .5750 |
| **Our Framework(IBEN)** | | .5516 |
| **Edited Headlines** | | |
| Glove | Bi-GRU | .6291 |
| Glove+FastText | Bi-GRU | .6212 |
| Glove+FastText+GoogleNews | CNN | .6057 |
| BERT(using layers(1,2,23,24)) | Bi-GRU | .5879 |
| BERT(using layers(1,2,3,4,24,23,22,21)) | Bi-GRU | .5701 |
| **Original Headlines** | | |
| Glove | Bi-GRU | .6311 |
| Glove+FastText | Bi-GRU | .6232 |
| Glove+FastText+GoogleNews | CNN | .6370 |
| BERT(using layers(1,2,23,24)) | Bi-GRU | .6194 |
| BERT(using layers(1,2,3,4,24,23,22,21)) | Bi-GRU | .6045 |

There are some limitations of our system that we observed during experiments. Our system is being harmed by a lack of knowledge of cultural metaphors and sarcasm. The model is having trouble understanding the negative influence of negative emotions on humour, as well as the blurry lines between sarcastic and ironic humour.

## 5. CONCLUSION

In this paper, we proposed a framework (IBEN) to detect the level of funniness in written sentences. In our architecture, we used a combination of deep learning techniques such as multikernel convolution, Bidirectional GRU, and BERT. The integration of the BERT and external embeddings with Bi-GRU and CNN models provides great understanding of sentence. The results show the performance of our framework. The main contribution of our unified framework is to learn contextual information effectively which in turn improved humour detection performance. Despite the fact that we obtained competitive results, our approach still has a lot of space for improvement.

In the future, we have a plan to focus on specific labelled forms of humour, such as incongruity, sarcasm, irony, puns, and superiority. This could help to better understand how different modelling strategies can identify different root causes of humour.

## ACKNOWLEDGEMENTS

This work was partially supported by JSPS KAKENHI Grant Number 17H01746.

## REFERENCES

[1] R. Paredes, J. S. Cardoso, and X. M. Pardo, "Pattern recognition and image analysis: 7th Iberian conference, IbPRIA 2015 Santiago de Compostela, Spain, june 17–19, 2015 proceedings," Lect. Notes Comput. Sci. (including Subser. Lect. Notes Artif. Intell. Lect. Notes Bioinformatics), vol. 9117, 2015, doi: 10.1007/978-3-319-19390-8.

[2] J. Mao and W. Liu, "A BERT-based approach for automatic humor detection and scoring," CEUR Workshop Proc., vol. 2421, no. September 2019, pp. 197–202, 2019.

[3] A. Reyes, P. Rosso, and D. Buscaldi, "From humor recognition to irony detection: The figurative language of social media," Data Knowl. Eng., vol. 74, pp. 1–12, 2012, doi: 10.1016/j.datak.2012.02.005.

[4] X. Yan and T. Pedersen, "Who's to say what's funny? A computer using Language Models and Deep Learning, That's Who!," arXiv, 2017.

[5] A. Wen et al., "Desiderata for delivering NLP to accelerate healthcare AI advancement and a Mayo Clinic NLP-as-a-service implementation," npj Digit. Med., vol. 2, no. 1, pp. 1–7, 2019, doi: 10.1038/s41746-019-0208-8.

[6] M. Abdullah and S. Shaikh, "TeamUNCC at SemEval-2018 Task 1: Emotion Detection in English and Arabic Tweets using Deep Learning," pp. 350–357, 2018, doi: 10.18653/v1/s181053.

[7] M. Zhou, N. Duan, S. Liu, and H. Y. Shum, "Progress in Neural NLP: Modeling, Learning, and Reasoning," Engineering, vol. 6, no. 3, pp. 275–290, 2020, doi: 10.1016/j.eng.2019.12.014.

[8] R. A. Martin, N. A. Kuiper, L. J. Olinger, and K. A. Dance, "Humor, coping with stress, selfconcept, and psychological well-being1," Humor, vol. 6, no. 1, pp. 89–104, 1993, doi: 10.1515/humr.1993.6.1.89.

[9] F. Barbieri and H. Saggion, "Automatic detection of irony and humour in Twitter," Proc. 5th Int. Conf. Comput. Creat. ICCC 2014, no. 1975, 2014.

[10] N. Hossain, J. Krumm, and M. Gamon, "'President Vows to Cut <Taxes> Hair': Dataset and Analysis of Creative Text Editing for Humorous Headlines," no. iv, pp. 133–142, 2019, doi: 10.18653/v1/n19-1012.

[11] A. Purandare and D. Litman, "Humor : Prosody Analysis and Automatic Recognition for F * R * I * E * N * D * S *. Humor : Prosody Analysis and Automatic Recognition for F * R * I * E * N * D * S *," no. January 2006, 2015.

[12] J. M. Taylor and L. J. Mazlack, "Computationally Recognizing Wordplay in Jokes Theories of Humor," no. 1991, 2000.

[13] L. Alfredo and L. De Oliveira, "Sequential Convolutional Architectures for Multi-Sentence Text Classification CS224N - Final Project Report," Nlp.Stanford.Edu, 2014, [Online]. Available: http://nlp.stanford.edu/courses/cs224n/2015/reports/7.pdf.

[14] B. Farzin, P. Czapla, and J. Howard, "Applying a pre-trained language model to Spanish twitter humor prediction," CEUR Workshop Proc., vol. 2421, no. June, pp. 172–179, 2019.

[15] S. Castro, L. Chiruzzo, and A. Rosá, "Overview of the HAHA Task: Humor analysis based on human annotation at IberEval 2018," CEUR Workshop Proc., vol. 2150, pp. 187–194, 2018.

[16] R. Ortega-Bueno, C. E. Muñiz-Cuza, J. E. Medina Pagola, and P. Rosso, "UO UPV: Deep linguistic humor detection in Spanish social media," CEUR Workshop Proc., vol. 2150, no. August, pp. 203–213, 2018.

[17] A. Garain, "Humor analysis based on human annotation (HAHA)-2019: Humor analysis at tweet level using deep learning," CEUR Workshop Proc., vol. 2421, no. September, pp. 191–196, 2019.

[18] V. Blinov, V. Bolotova-Baranova, and P. Braslavski, "Large dataset and language model funtuning for humor recognition," ACL 2019 - 57th Annu. Meet. Assoc. Comput. Linguist. Proc. Conf., pp. 4027–4032, 2020, doi: 10.18653/v1/p19-1394.

[19] M. Khodak, N. Saunshi, and K. Vodrahalli, "A large self-annotated corpus for sarcasm," arXiv, pp. 641–646, 2017.

[20] N. Hossain, J. Krumm, M. Gamon, H. Kautz, and M. Corporation, "SemEval-2020 Task 7: Assessing Humor in Edited News Headlines," no. 2019, 2020.

[21] J. Devlin, M. W. Chang, K. Lee, and K. Toutanova, "BERT: Pre-training of deep bidirectional transformers for language understanding," NAACL HLT 2019 - 2019 Conf. North Am. Chapter Assoc. Comput. Linguist. Hum. Lang. Technol. - Proc. Conf., vol. 1, no. Mlm, pp. 4171–4186, 2019.

[22] A. Vaswani et al., "Attention is all you need," Adv. Neural Inf. Process. Syst., vol. 2017December, no. Nips, pp. 5999–6009, 2017.

[23] J. Alammar, "The Illustrated BERT , ELMo , and co . ( How NLP Cracked Transfer Learning )," Blog, 2018.
[24] S. Takase, N. Okazaki, and K. Inui, "Learning to compose distributed representations of relational patterns," Trans. Japanese Soc. Artif. Intell., vol. 32, no. 4, pp. 1–11, 2017, doi: 10.1527/tjsai.DG96.
[25] L. Chen and C. M. Lee, "Predicting Audience's Laughter During Presentations Using Convolutional Neural Network," no. c, pp. 86–90, 2018, doi: 10.18653/v1/w17-5009.
[26] P. Association for Computational Linguistics, E. Grave, A. Joulin, and T. Mikolov, "Transactions of the Association for Computational Linguistics.," Trans. Assoc. Comput. Linguist., vol. 5, pp. 135–146, 2017, [Online]. Available: https://transacl.org/ojs/index.php/tacl/article/view/999.
[27] R. Cascade-correlation and N. S. Chunking, "2 PREVIOUS WORK," vol. 9, no. 8, pp. 1–32, 1997.
[28] J. Chung, "Gated Recurrent Neural Networks on Sequence Modeling arXiv : 1412 . 3555v1 [ cs . NE ] 11 Dec 2014," pp. 1–9.
[29] Z. Wang and D. P. T. Nguyen, "Salary Prediction using Bidirectional Gated Recurrent Unit - Convolutional Neural N etworks Model," ICACSIS 2015 - 2015 Int. Conf. Adv. Comput. Sci. Inf. Syst. Proc., no. C, pp. 292–295, 2019.
[30] Y. Kim, "Convolutional Neural Networks for Sentence Classification," pp. 1746–1751, 2014.
[31] D. P. Kingma and J. L. Ba, "A : a m s o," pp. 1–15, 2015.
[32] N. Colic and F. Rinaldi, "Improving spacy dependency annotation and pos tagging web service using independent NER services," Genomics and Informatics, 2019. .
[33] E. Loper and S. Bird, "NLTK: The Natural Language Toolkit," pp. 63–70, 2002, doi: 10.3115/1225403.1225421.
[34] T. Mikolov, E. Grave, P. Bojanowski, C. Puhrsch, and A. Joulin, "Advances in pre-training distributed word representations," arXiv, no. 1, pp. 52–55, 2017.
[35] J. Pennington, R. Socher, and C. D. Manning, "GloVe: Global vectors for word representation," 2014, doi: 10.3115/v1/d14-1162.
[36] T. Mikolov, I. Sutskever, K. Chen, G. Corrado, and J. Dean, "Distributed representations ofwords and phrases and their compositionality," Adv. Neural Inf. Process. Syst., pp. 1–9, 2013.
[37] H. Xiao, "bert-as-service Documentation," 2020.
[38] M. Abadi et al., "TensorFlow: A system for large-scale machine learning," Proc. 12th USENIX Symp. Oper. Syst. Des. Implementation, OSDI 2016, pp. 265–283, 2016.
[39] J. D. Owens, M. Houston, D. Luebke, S. Green, J. E. Stone, and J. C. Phillips, "GPU computing," Proc. IEEE, vol. 96, no. 5, pp. 879–899, 2008, doi: 10.1109/JPROC.2008.917757.

**AUTHORS**

Rida Miraj received her B.Sc. (Engg.) degree from the University of Engineering and Technology (UET), Lahore, Pakistan in 2019. She is currently a Master student at the Toyohashi University of Technology, Toyohashi, Japan. Her research interests include data mining and, sentiment analysis, emotion analysis, and humour detection.

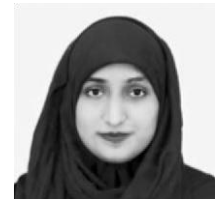

Masaki Aono received the BS and MS degrees from the Department of Information Science from the University of Tokyo, Tokyo, Japan, and the PhD degree from the Department of Computer Science at Rensselaer Polytechnic Institute, New York. He was with the IBM Tokyo Research Laboratory from 1984 to 2003. He is currently a professor at the Graduate School of Computer Science and Engineering Department, Toyohashi University of Technology. His research interests include text and data mining for massive streaming data, and information retrieval for multimedia including 2D images, videos, and 3D shape models. He is a member of the ACM and IEEE Computer Society. He has been a Japanese delegate of the ISO/IEC JTC1 SC24 Standard Committee since 1996.

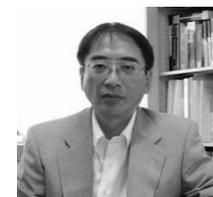